\documentclass[letterpaper, 10 pt, conference]{ieeeconf}  

\IEEEoverridecommandlockouts                              

\usepackage[usenames]{color}
\usepackage[svgnames]{xcolor}
\definecolor{DarkGreen}{rgb}{0,0.5,0}
\definecolor{DarkRed}{rgb}{0.75,0,0}
\definecolor{DarkOrange}{RGB}{255, 128, 0}
\usepackage{amsmath} 
\usepackage{nameref}
\usepackage{amssymb}
\usepackage{graphicx}
\usepackage{tikz}
\usetikzlibrary{shapes,snakes}
\usepackage{comment}
\usepackage[sort,compress]{cite}
\usepackage{url}
\usepackage[ruled,vlined,linesnumbered]{algorithm2e}
\usepackage{balance}
\usepackage[font=small]{caption}

\usepackage{array}

\newcommand{\bee}{\begin{enumerate} \itemsep -1pt \topsep -2pt}
\newcommand{\eee}{\end{enumerate}}

\usepackage[flushmargin]{footmisc}

\usepackage[colorlinks]{hyperref}
\hypersetup{citecolor=Black}
\hypersetup{linkcolor=Black}
\hypersetup{urlcolor=DarkBlue}
\usepackage{caption}

\usepackage{mathtools}
\usepackage[capitalize]{cleveref}
\crefformat{equation}{(#2#1#3)}
\Crefformat{equation}{Equation~(#2#1#3)}
\Crefname{equation}{Equation}{Equations}

\usepackage{bm}
\newcommand{\bb}[1]{\boldsymbol{{#1}}}
\newcommand{\vect}[1]{\boldsymbol{{#1}}}

\newcommand{\dvect}[1]{\boldsymbol{{\dot{#1}}}}
\newcommand{\ddvect}[1]{\boldsymbol{{\ddot{#1}}}}
\newcommand{\dddvect}[1]{\boldsymbol{{\dddot{#1}}}}

\usepackage[authormarkuptext=name,addedmarkup=bf,authormarkupposition=left]{changes}
\definechangesauthor[name={B.~L.}, color={blue}]{bl}
\setremarkmarkup{\bf(#2)}

\definecolor{MydarkRed}{RGB}{183,21,33}
\definecolor{MydarkGreen}{RGB}{59,143,50}
\definecolor{Myred}{RGB}{255,0,0}
\definecolor{MyorangeDarker}{RGB}{255,127,42}
\definecolor{MygreenDark}{RGB}{55,200,55}
\definecolor{MyorangeDark}{RGB}{255,212,42}
\definecolor{MyblueLight}{RGB}{85,221,255}
\definecolor{Myblue}{RGB}{0,0,255}
\newcommand{\tikzrectangle}[2][black,fill=red]{\tikz[baseline=0.0ex, line width=0.2mm]\draw[#1] [#1] (0,0) rectangle (0.2,0.2);}%

\newcommand{\tikzcircle}[2][red,fill=red]{\tikz[baseline=-0.5ex]\draw[#1,radius=#2] (0,0) circle ;}%

\newcommand{\subparagraph}{}
\usepackage{titlesec}
\titlespacing{\section}{8pt}{7pt}{6pt}

\renewcommand{\baselinestretch}{.98}
\let\emptyset\varnothing

\title{\LARGE \bf
FASTER: Fast and Safe Trajectory Planner for Flights\\ in Unknown Environments
}

\author{Jesus Tordesillas, Brett T. Lopez and Jonathan P. How
	\thanks{The authors are with the Aerospace Controls Laboratory, MIT, 77 Massachusetts Ave., Cambridge, MA, USA \tt\{jtorde, btlopez, jhow\}@mit.edu}
}
\begin{document}

\thispagestyle{empty}
\pagestyle{empty}
\maketitle
\begin{tikzpicture}[overlay, remember picture]
  \path (current page.north) ++(0.0,-1.0) node[draw = black] {Accepted for the 2019 IEEE/RSJ International Conference on Intelligent Robots and Systems (IROS), Macau, China};
\end{tikzpicture}
\vspace{-0.3cm}

\thispagestyle{empty}
\pagestyle{empty}

\begin{abstract}

High-speed trajectory planning through unknown environments requires algorithmic techniques that enable fast reaction times while maintaining safety as new information about the operating environment is obtained.
The requirement of computational tractability typically leads to optimization problems that do not include the obstacle constraints (collision checks are done on the solutions) or use a convex decomposition of the free space and then impose an ad-hoc time allocation scheme for each interval of the trajectory. 
Moreover, safety guarantees are usually obtained by having a local planner that plans a trajectory with a final ``stop" condition in the free-known space. 
However, these two decisions typically lead to slow and conservative trajectories.
We propose FASTER (Fast and Safe Trajectory Planner) to overcome these issues. 
FASTER obtains high-speed trajectories by enabling the local planner to optimize in both the free-known and unknown spaces. Safety guarantees are ensured by always having a feasible, safe back-up trajectory in the free-known space at the start of each replanning step. 
Furthermore, we present a Mixed Integer Quadratic Program formulation in which the solver can choose the trajectory interval allocation, and where a time allocation heuristic is computed efficiently using the result of the previous replanning iteration. 
This proposed algorithm is tested extensively both in simulation and in real hardware, showing agile flights in unknown cluttered environments with velocities up to $3.6$ m/s.

\end{abstract}

\section*{Supplementary material}

\textbf{Code}: FASTER (\href{https://github.com/mit-acl/faster}{https://github.com/mit-acl/faster}) and simulation worlds (\href{https://github.com/jtorde}{https://github.com/jtorde})

\textbf{Video}: \href{https://youtu.be/gwV0YRs5IWs}{https://youtu.be/gwV0YRs5IWs}.

\section{INTRODUCTION}

\begin{figure}[t]
	\centering
	\includegraphics[width=\columnwidth]{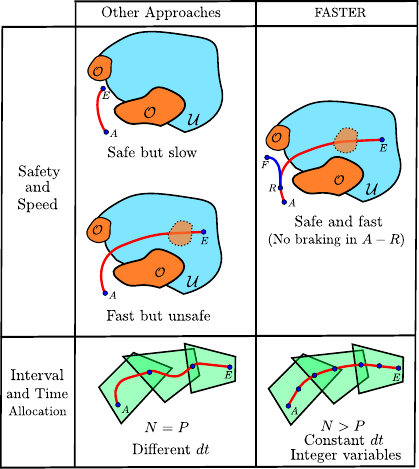}
	\caption{Contributions of this work. }
	\label{fig:contributions}
\end{figure} 

Navigating through unknown environments entails repeatedly generating collision-free, dynamically feasible trajectories that are executed over a finite horizon.
Similar to that in the Model Predictive Control (MPC) literature, safety is guaranteed by ensuring a feasible solution exists indefinitely.
If we consider $\mathbb{R}^3=\mathcal{O}\cup\mathcal{F}\cup\mathcal{U} $ where $\mathcal{F}$, $\mathcal{O}$, $\mathcal{U}$ are disjoint sets denoting free-known, occupied-known, and unknown space respectively, safety is guaranteed by constructing trajectories that are entirely contained in $\mathcal{F}$ with a final stop condition.
This can be achieved by generating motion primitives that do not intersect $\mathcal{O} \cup \mathcal{U}$ \cite{mueller2015computationally,florence2016integrated,lopez2017aggressive3D,lopez2017aggressivelimitedFOV}, or by constructing a convex representation of $\mathcal{F}$ to be used in an optimization \cite{liu2017planning, deits2015efficient, preiss2017trajectory}. 
However, both approaches lead to slow trajectories in scenarios where $\mathcal{F}$ is small compared to $\mathcal{U} \cup \mathcal{O}$.
This paper presents an optimization-based approach that reduces the aforementioned limitations by solving for \textit{two} optimal trajectories at every planning step (see Fig. \ref{fig:contributions}): one in  $\mathcal{U}\cup\mathcal{F}$, and another one in $\mathcal{F}$.


Decomposing the free space into $P$ overlapping polyhedra along a path connecting a start $A$ to goal $E$ location (see Fig. \ref{fig:contributions}), the usual approach is to divide the total trajectory into $N=P$ intervals \cite{liu2017planning}. On one hand, this simplifies the problem because no integer variables are needed, as each interval is forced to be in one specific polyhedron. On the other hand, the time allocation problem becomes much harder, as there are $N$ different $dt_n$ (time allocated for each interval $n$). The trajectory is also more conservative since the optimizer is only allowed to move the end points of each interval of the trajectory in the overlapping areas. To overcome these two problems, we propose the use of the same $dt$ for all the intervals, and use $N>P$ intervals, encoding the optimization problem as a Mixed Integer Quadratic Program (MIQP). Moreover, and as the minimum feasible $dt$ depends depends on the state of the UAV and on the specific shape of $\mathcal{F}$ and $\mathcal{U}$ at a specific replanning step, we also propose an efficient way to compute a heuristic of this $dt$ using the result obtained in the previous replanning iteration.

In summary, this work has the following contributions:
\begin{itemize}
    \item A framework that ensures feasibility of the entire collision avoidance algorithm and guarantees safety without reducing the nominal flight speed by allowing the local planner to plan in $\mathcal{F} \cup \mathcal{U}$  while always having a safe trajectory in $\mathcal{F}$.
	\item Reduced conservatism of the MIQP formulation for the interval and time allocation problem of the flight trajectories compared to prior work.
	\item Extension of our previous work \cite{tordesillas2018real}, (where we considered the interaction between the global and local planners) by proposing a way to compute very cheaply a heuristic of the cost-to-go needed by the local planner to decide which direction is the best one to optimize towards.
	\item Simulation and hardware experiments showing agile flights in completely unknown cluttered environments, with velocities up to $3.6$ m/s.

\end{itemize}

\section{RELATED WORK}
Trajectory planning strategies for UAVs can be classified according to the operating space of the local planner and the specific formulation of the optimization problem. 

With regard to the \textit{\textbf{planning space}} of the local planner, several approaches have been developed. 
One approach is to use only the most recent perception data \cite{lopez2017aggressivelimitedFOV, lopez2017aggressive3D, florence2016integrated}, which requires the desired trajectory to remain within the perception sensor field of view.
An alternative strategy is to create and plan trajectories in a map of the environment using a history of perception data. 
Within this second category, some works \cite{schouwenaars2002safe,tordesillas2018real,oleynikova2018safe} limit the local planner to generate trajectories only in \textit{free-known space} ($\mathcal{F}$ in \cref{fig:contributions}), which guarantees safety if the local planner has a final stop condition. 
However, limiting the planner to operating in free-known space and enforcing a terminal stopping condition can lead to conservative, slow trajectories (especially when much of the world is unknown).  
While allowing the local planner to optimize in both the free-known and unknown space ($\mathcal{F}\cup\mathcal{U}$), higher speeds can be obtained but with no guarantees that the trajectory is safe or will remain feasible. 

As far as the \textit{\textbf{optimization formulation}} is concerned, two approaches can be highlighted. The first does not include the obstacles in the optimization problem \cite{tordesillas2018real}, leading to a closed-form solution for the trajectory \cite{mueller2015computationally,lopez2017aggressive3D,lopez2017aggressivelimitedFOV} or in general to very small computation times \cite{tordesillas2018real}. 
The computation time for these approaches are very low since obstacles are not explicitly considered in the trajectory generation.
This enables multiple candidate trajectories to be generated (via sampling) and evaluated for collisions (using nearest-neighbor search) at each planning stage.
While these approaches are computationally efficient, they are unable to construct sophisticated maneuvers due to the discretization of the candidate trajectories, leading to slower trajectories in cluttered environments.
The second approach is to include obstacle constraints directly in the optimization.
This is usually done describing the free space by a set of  overlapping polyhedra (also known as convex decomposition) \cite{liu2017planning, deits2015efficient, preiss2017trajectory}. The trajectory can then be parameterized by a sequence of third (or higher)-degree polynomials.
B\'{e}zier Curves \cite{preiss2017trajectory}, \cite{sahingoz2014generation} or the sum-of-squares condition \cite{deits2015efficient, landry2016aggressive} can be used to guarantee that the trajectory remains in the overlapping polyhedra. 
Subsequently, there will be an interval (which polytope each polynomial is in) and a time allocation (how much time is assigned to each interval) problem. For the \textbf{interval allocation}, a typical solution is to use the same number of trajectory segments as number of polyhedra, and then each polynomial segment is forced to be inside its corresponding polyhedron \cite{preiss2017trajectory}. However, this can be very restrictive since the solver only has the freedom to select where the two endpoint points of each interval are placed in the overlapping regions. 
Another option, but with higher computation times, is to use binary decision variables \cite{landry2016aggressive, deits2015efficient} to allow the solver to choose the specific interval allocation. 
For the \textbf{time allocation}, one can either use a fixed time allocation \cite{liu2017planning} or formulate a bi-level optimization to find the times \cite{richter2016polynomial}, \cite{preiss2017trajectory}.
However, the first approach can be very conservative and can cause infeasibility in the optimization problem while the seconds leads to longer replanning times. 

\section{FAST AND SAFE TRAJECTORY PLANNER}
\subsection{Planning}

The Fast and Safe Trajectory Planner (FASTER) uses hierarchical architecture where a long-horizon global planner guides a short-horizon local planner to a desired goal location. 
The global planner used in this work is Jump Point Search (JPS). JPS finds the shortest piecewise linear path between two points in a 3D uniformly-weighted voxel grid, guaranteeing optimality and completeness but running an order of magnitude faster than A* \cite{harabor2011online}, \cite{liu2017planning}.

For the local planner, we distinguish these three different jerk-controlled trajectories (some of the points will be precisely defined later, see Fig. \ref{fig:plan2}):
\begin{itemize}

	\item \textbf{\textcolor{red}{Whole Trajectory}:} This trajectory goes from a start location $A$ to goal location $E$, and it is contained in $\mathcal{F}\cup\mathcal{U}$. It has a final stop condition.
	\item \textbf{\textcolor{blue}{Safe Trajectory}:} It goes from $R$ to $F$, where $R$ is a point in the Whole Trajectory, and $F$ is any point inside the polyhedra obtained by doing a convex decomposition of $\mathcal{F}$. It is completely contained in $\mathcal{F}$ (free-known space), and it has also a final stop condition to guarantee safety.
	\item \textbf{\textcolor{ForestGreen}{Committed Trajectory}:} This trajectory consists of two pieces: The first part is the interval $A \rightarrow R$ of the Whole Trajectory. The second part is the Safe Trajectory. It is also guaranteed to be inside $\mathcal{F}$ (see explanation below). This trajectory is the one that the UAV will execute in case no feasible solutions are found in the next replanning steps. 

\end{itemize}

The quadrotor is modeled using triple integrator dynamics with state vector
$
\mathbf{x}^T = \left[ \vect{x}^T ~ \dvect{x}^T ~ \ddvect{x}^T ~ \right] = \left[\vect{x}^T ~\vect{v}^T ~\vect{a}^T\right]
$
and control input $\mathbf{u} = \dddvect{x} = \vect{j}$ (where $\vect{x}$, $\vect{v}$, $\vect{a}$, and $\vect{j}$ are the vehicle's position, velocity, acceleration, and jerk, respectively).

Let $n=0:N-1$ denote the specific interval of the trajectory and $p=0:P-1$ the specific polyhedron. If $\bb{j}(t)$ is constrained to be constant in each interval $n=0:N-1$, then the whole trajectory will be a spline consisting of third degree polynomials. Matching the cubic form of the position for each interval $$\boldsymbol{x}_{n}(\tau)=\boldsymbol{a}_{n}\tau^{3}+\boldsymbol{b}_{n}\tau^{2}+\boldsymbol{c}_{n}\tau+\boldsymbol{d}_{n},\; \; \tau\in[0,dt]$$ with the expression of a cubic B\'{e}zier curve
$$\boldsymbol{x}_{n}(\tau)=\sum_{j=0}^{3}\left(\begin{array}{c}
3\\
j
\end{array}\right)\left(1-\frac{\tau}{dt}\right)^{3-j}\left(\frac{\tau}{dt}\right)^{j}\boldsymbol{r}_{nj},\;\; \tau\in[0,dt],$$
we can solve for the four control points  $\bb{r}_{nj}$ $(j=0:3)$ associated with each interval $n$:

\begin{align} & \boldsymbol{r}_{n0}=\boldsymbol{d}_{n}, \qquad 
	\boldsymbol{r}_{n1}=\frac{\boldsymbol{c}_{n}dt+3\boldsymbol{d}_{n}}{3} & \nonumber\\
	& \boldsymbol{r}_{n2}=\frac{\boldsymbol{b}_{n}dt^{2}+2\boldsymbol{c}_{n}dt+3\boldsymbol{d}_{n}}{3} & \nonumber\\
	& \boldsymbol{r}_{n3}=\boldsymbol{a}_{n}dt^{3}+\boldsymbol{b}_{n}dt^{2}+\boldsymbol{c}_{n}dt+\boldsymbol{d}_{n}  & \nonumber
\end{align}

Let us denote the sequence of $P$ overlapping polyhedra as $\{(\bb{A}_p , \bb{c}_p)\},\;p=0:P-1$, and introduce binary variables $b_{np}$ ($P$ variables for each interval $n=0:N-1$). As a B\'{e}zier curve is contained in the convex hull of its control points, we can ensure that the whole trajectory will be inside this convex corridor by forcing that all the control points are in the same polyhedron \cite{preiss2017trajectory, sahingoz2014generation} with the constraint $[b_{np}=1\implies \boldsymbol{r}_{nj} \in \text{polyhedron $p$} \;\; \forall j]$, and at least in one polyhedron with the constraint $\sum_{p=0}^{P-1}b_{np}\ge1$. The optimizer is free to choose in which polyhedron exactly.
The complete MIQP solved in each replanning step (using Gurobi, \cite{gurobi}) for both the Safe and the Whole trajectories is this one:
\begin{align}\min_{\boldsymbol{j}_{n}, b_{np}} & \sum_{n=0}^{N-1}\left\Vert \mathbf{j}_{n}\right\Vert ^{2}\label{eq:MIQP}\\
	\textrm{s.t. } & \mathbf{x}_{0}(0)=\mathbf{x}_{init}\nonumber\\
	& \mathbf{x}_{N-1}(dt)=\mathbf{x}_{final}\nonumber\\
	& \boldsymbol{x}_{n}(\tau)=\boldsymbol{a}_{n}\tau^{3}+\boldsymbol{b}_{n}\tau^{2}+\boldsymbol{c}_{n}\tau+\boldsymbol{d}_{n}\;\forall n,\forall\tau\in[0,dt] & \nonumber\\
	& \boldsymbol{v}_{n}(\tau)=\dot{\boldsymbol{x}}_{n}(\tau)\;\forall n,\forall\tau\in[0,dt]\nonumber\\
	& \boldsymbol{a}_{n}(\tau)=\dot{\boldsymbol{v}}_{n}(\tau)\;\forall n,\forall\tau\in[0,dt]\nonumber\\
	& \mathbf{j}_{n}=6\boldsymbol{a}_{n}(0)\;\forall n & \nonumber\\
	& b_{np}=1\implies\left\{ \begin{array}{c} 
		\boldsymbol{A}_{p}\boldsymbol{r}_{n0}\le\boldsymbol{c}_{p}\\
		\boldsymbol{A}_{p}\boldsymbol{r}_{n1}\le\boldsymbol{c}_{p}\\
		\boldsymbol{A}_{p}\boldsymbol{r}_{n2}\le\boldsymbol{c}_{p}\\
		\boldsymbol{A}_{p}\boldsymbol{r}_{n3}\le\boldsymbol{c}_{p}
	\end{array}\right.\quad\forall n,\forall p & \nonumber\\
	& \sum_{p=0}^{P-1}b_{np}\ge1\quad\forall n &\nonumber\\
	& b_{np}\in\{0,1\}\quad\forall n,\forall p & \nonumber\\
	& \mathbf{x}_{n+1}(0)=\mathbf{x}_{n}(dt)\quad n=0:N-2 & \nonumber\\
	& \left\Vert \boldsymbol{v}_{n}(0)\right\Vert _{\infty}\le v_{max}, \qquad\quad\forall n & \nonumber\\
	& \left\Vert \boldsymbol{a}_{n}(0)\right\Vert _{\infty}\le a_{max},\qquad\quad\forall n & \nonumber\\
	& \left\Vert \boldsymbol{j}_{n}\right\Vert _{\infty}\le j_{max},\qquad\quad\quad\;\;\forall n & \nonumber
\end{align}

\begin{figure}[]
	\centering
	\includegraphics[width=\columnwidth]{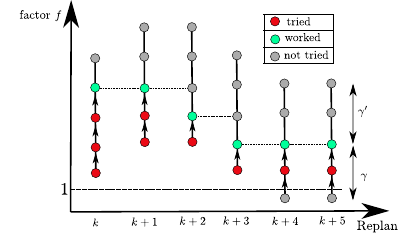}
	\caption{Dynamic adaptation of the factor used to compute the heuristic of the time allocation per interval ($dt$): For iteration $k$, the range of factors used is taken around the factor that worked in the iteration $k-1$. }
	\label{fig:dynamic_adaptation_factor}
	\vspace*{-.15in}
\end{figure}

\begin{figure}[]
	\centering
	\includegraphics[width=\columnwidth]{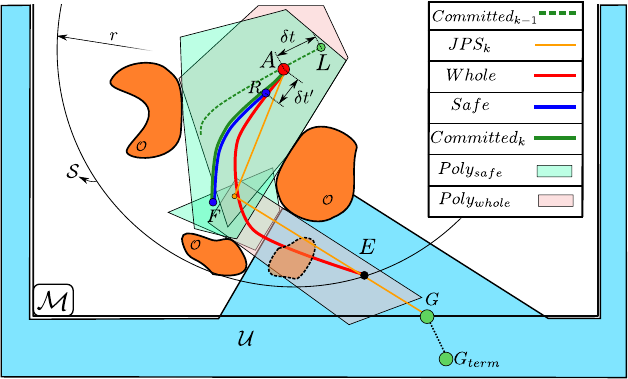}
	\caption{Illustration for Alg.\ref{algo: myalgorithm_iros}. $\mathcal{U}$ is the unknown space  (\tikzrectangle[black,fill=MyblueLight]{10pt}), and $\mathcal{O}$ are the known obstacles  (\tikzrectangle[black,fill=DarkOrange]{10pt}) . One unknown obstacle is shown with dotted line.}
	\label{fig:plan2}
	\vspace*{-0.25in}
\end{figure}

In the optimization problem above, $dt$ (same for every interval $n$) is computed as 
\begin{equation}\label{eq: find_dt}
dt=f\cdot\max\{T_{v_x},T_{v_y}, T_{v_z}, T_{a_x},T_{a_y},T_{a_z},T_{j_x},T_{j_y},T_{j_z}\}/N 
\end{equation}
where $T_{v_i}$, $T_{a_i}$, $T_{j_i}$ are solution of the constant-input motions in each axis $i={x,y,z}$ by applying $v_{max}$, $a_{max}$ and $j_{max}$ respectively. $f \ge 1$ is a factor that is obtained according to the solution of the previous replanning step (see Fig. \ref{fig:dynamic_adaptation_factor}): The optimizer will try values of $f$ (in increasing order) in the interval  $[f_{worked,k-1}-\gamma,f_{worked,k-1}+\gamma']$ until the problem converges. Here, $f_{worked,k-1}$ is the factor that made the problem feasible in the previous replanning step. Note that, if $f=1$, then $dt$ is a lower bound on the minimum time per interval required for the problem to be feasible.

\newcommand{\argmin}{\mathop{\mbox{argmin}}}
\newcommand{\argmax}{\mathop{\mbox{argmax}}}

\begin{algorithm}[]
	\footnotesize
	
	\DontPrintSemicolon
	\KwData{Current Position of the UAV $L$, $Committed_{k-1}$, $JPS_{k-1}$, $G_{term}$, $\mathcal{O}$, $\mathcal{F}$, $\mathcal{U}$,  $r$   }
	
	\SetKwFunction{FMain}{\textbf{Replan}}
	\SetKwProg{Pn}{Function}{:}{\KwRet}
	\Pn{\FMain{}}{
		
		$k\leftarrow k+1 $, $\delta t\leftarrow\alpha \Delta t_{k-1}$, $\delta t'\leftarrow\beta \Delta t_{k-1}$  \;
Choose point $A$ in $\textcolor{ForestGreen}{Committed}_{k-1}$ with offset $\delta t$  from $L$\;
$G\leftarrow $ Projection of $G_{term}$ into map $\mathcal{M}$ \label{projection}\;
$JPS_a\leftarrow$ Run JPS $A \rightarrow G$\label{points_end}\;

$C\leftarrow JPS_a \cap \mathcal{S}$
\label{cost_start}\;
$JPS_b\leftarrow$ Modified $JPS_{k-1}$ such that $JPS_{k-1} \cap \mathcal{O}=\emptyset$\;
$D\leftarrow JPS_{b} \cap \mathcal{S}$\;
$dt_{a} \leftarrow$  Lower bound on dt $A \rightarrow C$\;
$dt_{b} \leftarrow$  Lower bound on dt $A \rightarrow D$\;
$J_a=N \cdot dt_a+\frac{ \left\Vert JPS_a(C \rightarrow G) \right\Vert }{v_{max}}$ \;
$J_b=N \cdot dt_b+ \frac{ \left\Vert JPS_b(D \rightarrow G) \right\Vert }{v_{max}}$\;
$JPS_k\leftarrow \underset{\{JPS_{a,}JPS_{b}\}}{\argmin}\{J_{a},J_{b}\}$\;

$JPS_k\leftarrow JPS_a$ \label{cost_end}

$JPS_{in} \leftarrow $ Part of $JPS_k$ inside $\mathcal{S}$\label{whole_start}\;
$Poly_{whole} \leftarrow $ Convex Decomposition in $\mathcal{U}\cup\mathcal{F}$ using $JPS_{in}$\;
$f_{whole} \leftarrow [f_{whole,k-1}-\gamma,  f_{whole,k-1}+\gamma' ] $\;
$\textcolor{red}{Whole} \leftarrow$ MIQP in $Poly_{whole}$ from $A$ to $G$  using $f_{whole}$\label{whole_end}\;
$JPS_{in,known} \leftarrow $ Part of $JPS_{in}$ in $\mathcal{F}$\label{safe_start}\;
$Poly_{safe} \leftarrow $Convex Decomposition in  $\mathcal{F}$ using $JPS_{in,known}$\;
$f_{safe} \leftarrow [f_{safe,k-1} -\gamma, f_{safe,k-1} +\gamma' ] $\;
\textcolor{blue}{$Safe$} $\leftarrow$ MIQP in $Poly_{safe}$ from $R$ to $F$ using $f_{safe}$\label{safe_end}\;

$\textcolor{ForestGreen}{Committed}_k \leftarrow \textcolor{red}{Whole}_{A \rightarrow R}\cup \textcolor{blue}{Safe}\label{committed}$\;
$f_{whole,k}\leftarrow$ Factor that worked for \textcolor{red}{$Whole$}\;
$f_{safe,k}\leftarrow$ Factor that worked for \textcolor{blue}{$Safe$}\;
$\Delta t_{k}\leftarrow$ Total replanning time\;

	}
	\normalsize
	\caption{FASTER \label{IR}}
	\label{algo: myalgorithm_iros}
\end{algorithm}

\begin{figure}[]
	\centering
	\includegraphics[width=1\columnwidth]{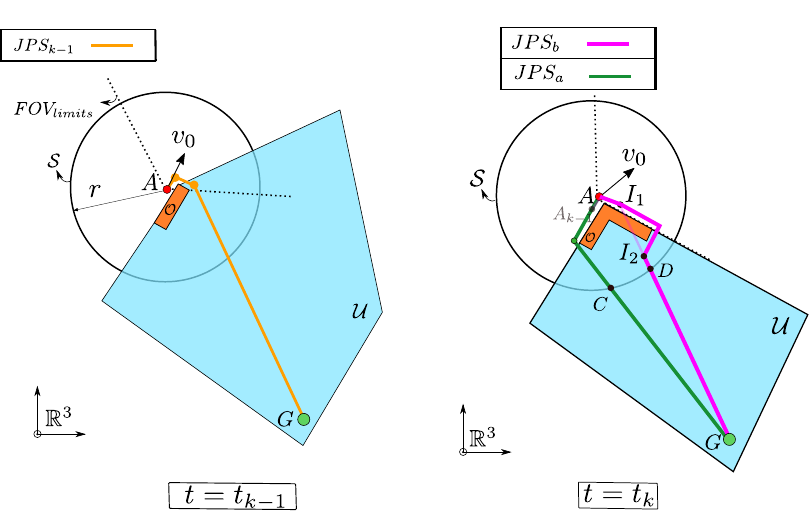}
	\caption[Choice of the direction to optimize]{ Choice of the direction to optimize. At $t=t_{k-1}$, the JPS solution chosen was $JPS_{k-1}$. At $t=t_{k}$, JPS is run again to obtain $JPS_a$, and $JPS_{k-1}$ is modified so that it does not collide with $\mathcal{O}$, obtaining $JPS_b$. A heuristic of the cost-to-go in each direction is computed, and the direction with the lowest cost is chosen as the one towards which the local planner will optimize.}
	\label{fig:planning_strategy1}
\end{figure}

Algorithm~\ref{algo: myalgorithm_iros} summarizes the full approach (see also Fig. \ref{fig:plan2}). Let $L$ be the current position of the UAV. The point $A$ is chosen in the Committed Trajectory of the previous replanning step with an offset $\delta t$ from $L$. This offset $\delta t$ is computed by multiplying the total time of the previous replanning step by $\alpha\ge1$ (typically $\alpha \approx 1.25$). The idea here is to dynamically change this offset to ensure that most of the times the solver is able to find the next solution in less than $\delta t$. Then, the final goal $G_{term}$ is projected into the sliding map $\mathcal{M}$ (centered on the UAV) in the direction $\overrightarrow{G_{term}A}$ to obtain the point $G$ (line \ref{projection}). Next, we run JPS from $A$ to $G$ (line \ref{points_end}) to obtain $JPS_a$.
 
The local planner then must decide if the current JPS solution should be used to guide the optimization (lines \ref{cost_start}-\ref{cost_end}). Instead of blindly trusting the last JPS solution ($JPS_a$) as the best direction for the local planner to optimize (note that JPS is a zero-order model, without dynamics encoded), we take into account the dynamics of the UAV in the following way: First of all, we modify the $JPS_{k-1}$ so that it does not collide with the new obstacles seen (Fig. \ref{fig:planning_strategy1}): we find the points $I_1$ and $I_2$ (first and last intersections of $JPS_{k-1}$ with $\mathcal{O}$) and run JPS three times, so  $A \rightarrow I_1$,  $I_1 \rightarrow I_2$ and  $I_2 \rightarrow I_G$. Hence, the modified version, denoted by $JPS_{b}$, will be the concatenation of these three paths. 
Then, we compute a lower bound on $dt$ using Eq. \ref{eq: find_dt} for both $A \rightarrow C$ and $A \rightarrow D$, where $C$ and $D$ are the intersections of the previous JPS paths with a sphere $\mathcal{S}$ of radius $r$ centered on $A$. Next, we find the cost-to-go associated with each direction by adding this $dt_a$ (or $dt_b$) and the time it would take the UAV to go from $C$ (or $D$) to $G$ following the JPS solution and flying at $v_{max}$. Finally, the one with lowest cost is chosen, and therefore $JPS_k\leftarrow \underset{\{JPS_{a,}JPS_{b}\}}{\argmin}\{J_{a},J_{b}\}$. This will be the direction towards which the local planner will optimize.
 
The \textcolor{red}{Whole Trajectory} (lines \ref{whole_start}-\ref{whole_end}) is obtained as follows. We do the convex decomposition \cite{liu2017planning} of $\mathcal{U} \cup \mathcal{F}$ around $JPS_{in}$, which is the part of $JPS_{k}$ that is inside the sphere $\mathcal{S}$. This gives a series of overlapping polyhedra that we denote as $Poly_{whole}$. Then, the MIQP in (\ref{eq:MIQP}) is solved using these polyhedral constraints to obtain the Whole Trajectory.
 
The \textcolor{blue}{Safe Trajectory} is computed as in lines \ref{safe_start}-\ref{safe_end}. First we choose the point $R$ along the Whole Trajectory with an offset $\delta t'$ from $A$ (this $\delta t'$ is computed by multiplying the previous replanning time by $\beta\ge1$), and run convex decomposition in $\mathcal{F}$ using the part of $JPS_{in}$ that is in $\mathcal{F}$, obtaining the polyhedra $Poly_{safe}$. Then, we solve the MIQP from $R$ to any point $F$ inside $Poly_{safe}$ (this point $F$ is chosen by the optimizer).
 
In both of the convex decompositions presented above, one polyhedron is created for each segment of the piecewise linear paths. To obtain a less conservative solution (i.e. bigger polyhedra), we first check the length of segments of the JPS path, creating more vertexes if this length exceeds certain threshold $l_{max}$. Moreover, we truncate the number of segments in the path to ensure that the number of polyhedra found does not exceed a threshold $P_{max}$. This helps reduce the computation times (see Sec. \ref{sec:results}).
 
Finally (line \ref{committed}), we compute the \textcolor{ForestGreen}{Committed Trajectory} by concatenating the piece $A \rightarrow R$ of the Whole Trajectory, and the Safe Trajectory. Note that in this algorithm we have run two \textit{decoupled} optimization problems per replanning step: (1) one for the Whole Trajectory, and (2) one for the Safe Trajectory. This ensures that the piece $A \rightarrow R$ is not influenced by the braking maneuver $R \rightarrow F$, and therefore guarantees a higher nominal speed on this first piece. The intervals $L \rightarrow A$ and $A \rightarrow R$ have been designed so that, with high probability, at least one replanning step can be solved within that interval. Moreover, to prevent the (very rare) cases where both $A$ and $R$ are in $\mathcal{F}$, but the piece $A-R$ is not, we check that piece $A-R$ against collision with $\mathcal{U}$. If any of the two optimizations in this algorithm fails, or the piece $A-R$ intersects $\mathcal{U}$, or the replanning step takes longer than $\delta t$, the UAV does not commit to a new trajectory in that replanning step, and continues executing the Committed Trajectory of the previous replanning step. Thus safety is guaranteed by construction: the UAV will only fly Committed Trajectories, which are always guaranteed to be in $\mathcal{F}$ with a terminal stopping condition. 
 
 \subsection{Mapping}
 For the mapping, we use a sliding map centered on the UAV that moves as the UAV flies. We fuse a depth map into the occupancy grid using the 3D Bresenham’s line algorithm for ray-tracing \cite{bresenham1965algorithm}, and  $\mathcal{O}$ and $\mathcal{U}$ are inflated by the radius of the UAV to ensure safety.

\section{RESULTS} \label{sec:results}

\subsection{Simulation}

\begin{figure}[t]
	\centering
	\includegraphics[width=\columnwidth]{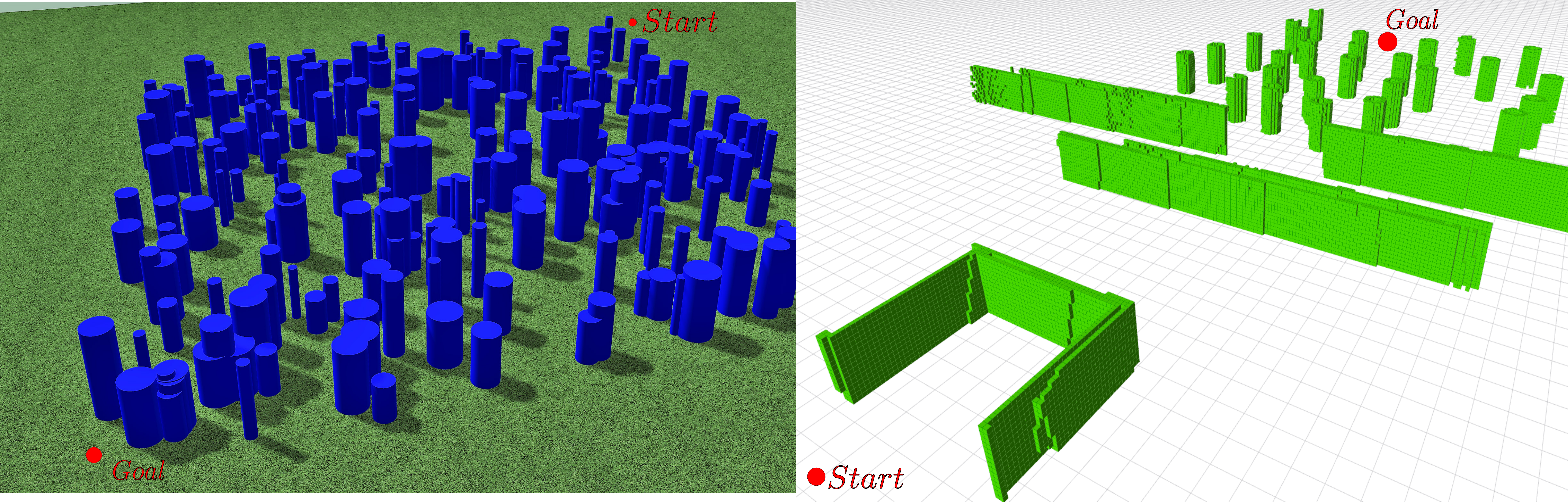}
	\caption[Forest and bugtrap environments used in the simulation]{Forest (left) and bugtrap (right) environments used in the simulation. The forest is $50$~$\times$~$50$~m, and the grid in the bugtrap environment is $1$~m~$\times$~$1$~m.}
	\label{fig:forest_and_bugtrap}
\end{figure}

We first test FASTER in 10 random forest environments with an obstacle density of $0.1$~obstacles/m$^2$ (see Fig.~\ref{fig:forest_and_bugtrap}) and compare the flight distances achieved against the following seven approaches: Incremental approach (no goal selection), random goal selection, optimistic RRT$^\star$ (unknown space = free), conservative RRT$^\star$ (unknown space = occupied), ``next-best-view" planner (NBVP) \cite{bircher2016receding}, Safe Local Exploration \cite{oleynikova2018safe}, (see \cite{oleynikova2018safe} for details of all these approaches), and Multi-Fidelity \cite{tordesillas2018real}. 

The results are shown in Table~\ref{tab:table_forest_distance}, which highlights that FASTER achieves a $8-51\%$ improvement in the distance. Completion times are compared in Table \ref{tab:table_forest_time} to our previous proposed algorithm \cite{tordesillas2018real} (time values are not available for all other algorithms in Table~\ref{tab:table_forest_distance}). FASTER achieves an improvement of $52\%$ in the completion time. The dynamic constraints imposed for the results of this table are (per axis) $v_{max}=5$ m/s, $a_{max}= 5$ m/s$^2$, and $j_{max}= 8$ m/s$^3$.

We also test FASTER using the bugtrap environment shown in Fig. \ref{fig:forest_and_bugtrap}, and obtain the results that appear on Table \ref{tab:table_bugtrap}. Both algorithms have a similar total distance, but FASTER achieves an improvement of $63\%$ on the total flight time. For both cases the dynamic constraints imposed are $v_{max}=10$ m/s, $a_{max}= 10$ m/s$^2$, and $j_{max}= 40$  m/s$^3$.

\begin{figure}[]
	\centering
	\includegraphics[width=1\columnwidth]{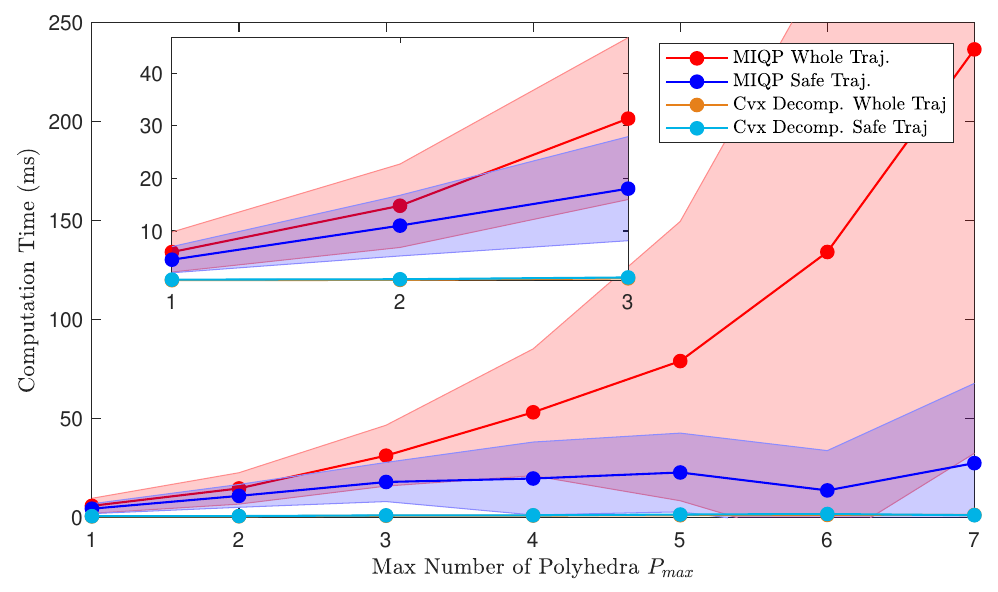}
	\caption{Timing breakdown for the MIQP and Convex Decomposition of the Whole Trajectory and the Safe Trajectory as a function of the maximum number of polyhedra $P_{max}$. Note that the times for the MIQPs include all the trials until convergence (with different factors $f$) in each replanning step. The shaded area is the 1-$\sigma$ interval, where $\sigma$ is the standard deviation. These results are from the forest simulation.}
	\label{fig:timing_all}
\end{figure} 

The timing breakdown of Alg.~\ref{algo: myalgorithm_iros} as a function of the maximum number of polyhedra $P_{max}$ is shown in Fig. \ref{fig:timing_all}. The number of intervals $N$ was 10 for the Whole Trajectory and 7 for the Safe Trajectory. Note that the runtime for the MIQP of the Safe Trajectory is approximately constant as a function of $P_{max}$. This is due to the fact that the Safe Trajectory is planned only in $\mathcal{F}$, and therefore most of the times $P < P_{max}$. For the simulations and hardware experiments presented in this paper, $P_{max} = 2-3$ was used. The runtimes for JPS as a function of the voxel size of the map for the forest simulation are available in Fig. 7 of \cite{tordesillas2018real}. All these timing breakdowns were measured using an Intel Core i7-7700HQ 2.8GHz Processor.

\begin{table}[t]
	\caption{\label{tab:table_forest_distance}Distances obtained in 10 random forest simulations. \iffalse Distance values are computed for cases that reach the goal.\fi Improvement percentages are computed for the minimum and the maximum of each column. Some results were provided by the authors of \cite{oleynikova2018safe}.}
	\vspace*{-.075in}
	\begin{tabular}{p{1.7cm} >{\centering\arraybackslash}p{1.5cm} >{\raggedleft\arraybackslash}p{0.75cm} >{\raggedleft\arraybackslash}p{0.75cm} >{\raggedleft\arraybackslash}p{0.75cm} >{\raggedleft\arraybackslash}p{0.75cm}}
		\hline
		\hline
		\multicolumn{1}{l}{\textbf{Method}} & \multicolumn{1}{l}{\textbf{Number of}}     & \multicolumn{4}{c}{\textbf{Distance (m)}}                                                            \\ \cline{3-6} 
		\multicolumn{1}{l}{}               & \textbf{Successes} & \textbf{Avg}  & \textbf{Std} & \multicolumn{1}{l}{\textbf{Max}} & \multicolumn{1}{l}{\textbf{Min}} \\ 
		Incremental                        & 0                  & -             & -            & -                                & -                                \\ 
		Rand. Goals                        & \textbf{10}        & 138.0         & 32.0         & 210.5                            & 105.6                            \\ 
		Opt. RRT$^\star$                   & 9                  & 105.3         & 10.3         & 126.4                            & 95.5                             \\ 
		Cons. RRT$^\star$                  & 9                  & 155.8         & 52.6         & 267.9                            & 106.2                            \\ 
		NBVP  \cite{bircher2016receding}   & 6                  & 159.3         & 45.6         & 246.9                            & 123.6                            \\ 
		SL Expl. \cite{oleynikova2018safe} & 8                  & 103.8         & 21.6         & 148.3                            & 86.6                             \\ 
		Mult-Fid \cite{tordesillas2018real}         &  \textbf{10}                 & 84.5          & 11.7         & 109.4                            & 73.2                             \\ 
		\textbf{FASTER}                      &  \textbf{10}                 & \textbf{77.6} & \textbf{5.9} & \textbf{88.0}                    & \textbf{70.7}                    \\ \hline
		\multicolumn{2}{l}{\hspace*{-.5em} \rule{0pt}{10pt} \textbf{Min/Max improvement (\%)}}                          & \textbf{8/51}              & \textbf{43/89} & \textbf{20/67} & \textbf{3/43}                                       \\ \hline \hline
	\end{tabular}
\vspace*{.2in}			
%
%
	\caption{\label{tab:table_forest_time} Comparison between \cite{tordesillas2018real} and FASTER of flight times in the forest simulation. Results are for 10 random forests. }
\vspace*{-.075in}
\begin{tabular}{p{1.5cm} >{\centering\arraybackslash}p{1.5cm} >{\raggedleft\arraybackslash}p{0.75cm} >{\raggedleft\arraybackslash}p{0.75cm} >{\raggedleft\arraybackslash}p{0.75cm} >{\raggedleft\arraybackslash}p{0.75cm}}
		\hline
		\hline
		\multicolumn{1}{l}{\textbf{Method}} &       & \multicolumn{4}{c}{\textbf{Time (s)}}                                                            \\ \cline{3-6} 
		\multicolumn{1}{l}{}               &   & \textbf{Avg}  & \textbf{Std} & \multicolumn{1}{l}{\textbf{ Max}} & \multicolumn{1}{l}{\textbf {Min}} \\ 
		Mult-Fid \cite{tordesillas2018real}         &                   & 61.2          & 16.8         & 92.5                             & 37.9                             \\ 
		\textbf{FASTER}                      &          & \textbf{29.2} & \textbf{4.2} & \textbf{36.8}                    & \textbf{21.6}                    \\ \hline
		\multicolumn{2}{l}{\hspace*{-.5em} \rule{0pt}{10pt} \textbf{Improvement (\%)}}                          & \textbf{52.3}              & \textbf{75.0} & \textbf{60.2} & \textbf{43.0}                                       \\ \hline \hline
	\end{tabular}
						
%
%
\vspace*{.2in}
\caption{\label{tab:table_bugtrap} Comparison between \cite{tordesillas2018real} and FASTER of flight distances and times in a bugtrap simulation.}
\vspace*{-.075in}
	\centering 
	\begin{tabular}{p{2cm} >{\centering\arraybackslash}p{2cm} >{\raggedleft\arraybackslash}p{1.75cm} >{\raggedleft\arraybackslash}p{1.75cm}  }
		\hline
		\hline

		\multicolumn{1}{l}{\textbf{Method}}               & \textbf{Distance (m)}  & \textbf{Time (s)} \\ 
		Mult-Fid \cite{tordesillas2018real}         & 56.8                   & 37.6             \\ 
		\textbf{FASTER}                      & \textbf{55.2}         & \textbf{13.8}    \\ \hline
		\multicolumn{1}{l}{\hspace*{-.8em} \rule{0pt}{10pt} \textbf{Improvement (\%)}}    & \textbf{2.8}              & \textbf{63.3}                                    \\ \hline \hline
	\end{tabular}
						
\end{table}

\begin{figure}[t]
	\centering
	\includegraphics[width=\columnwidth]{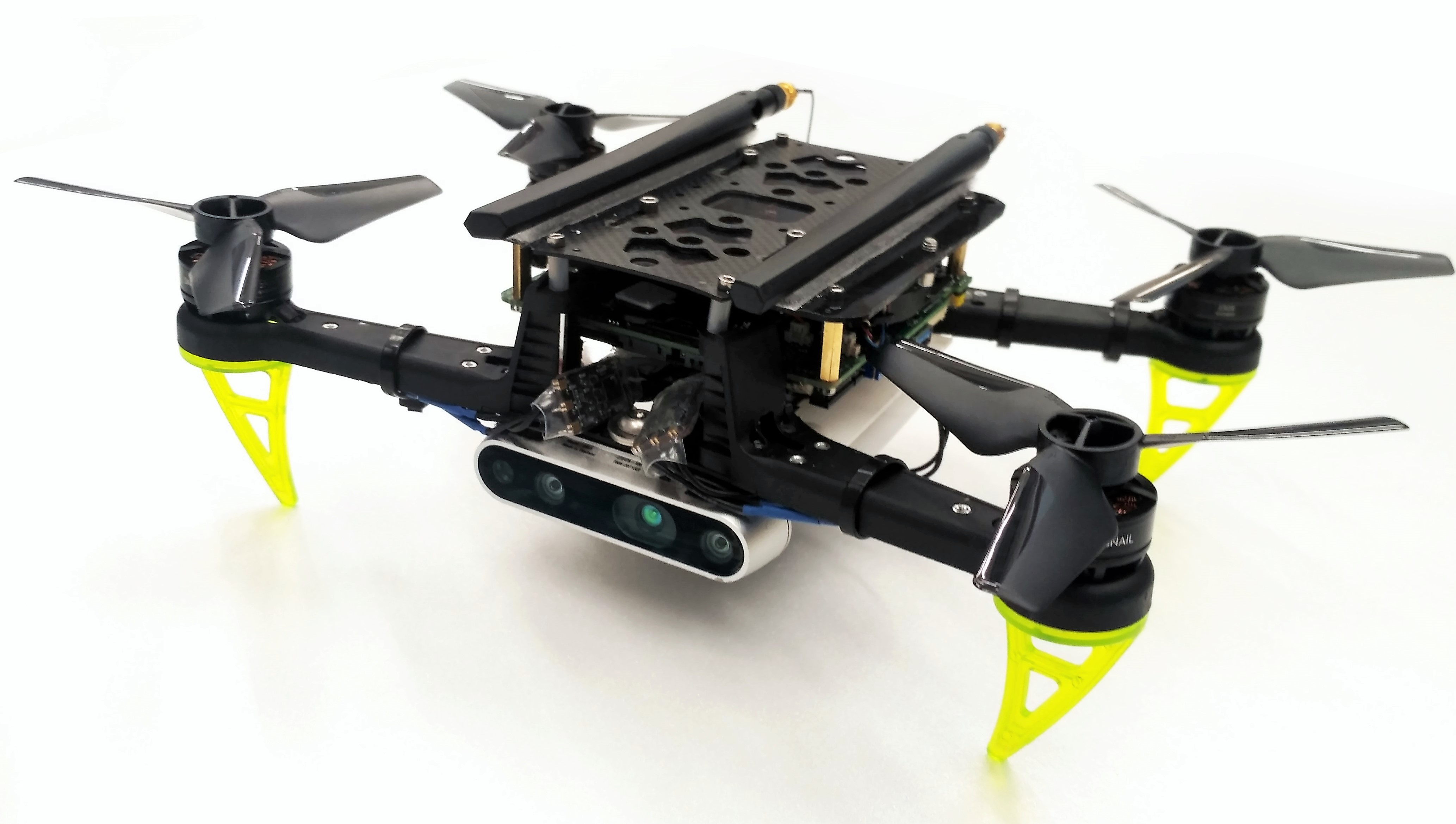}
	\caption{UAV used in the experiments. It is equipped with a Qualcomm\textsuperscript{\textregistered} SnapDragon Flight, an Intel\textsuperscript{\textregistered} NUC and an Intel\textsuperscript{\textregistered} RealSense Depth Camera D435.  }
	\label{fig:drone}
\end{figure}

\begin{figure*}[]
	\includegraphics[width=\textwidth]{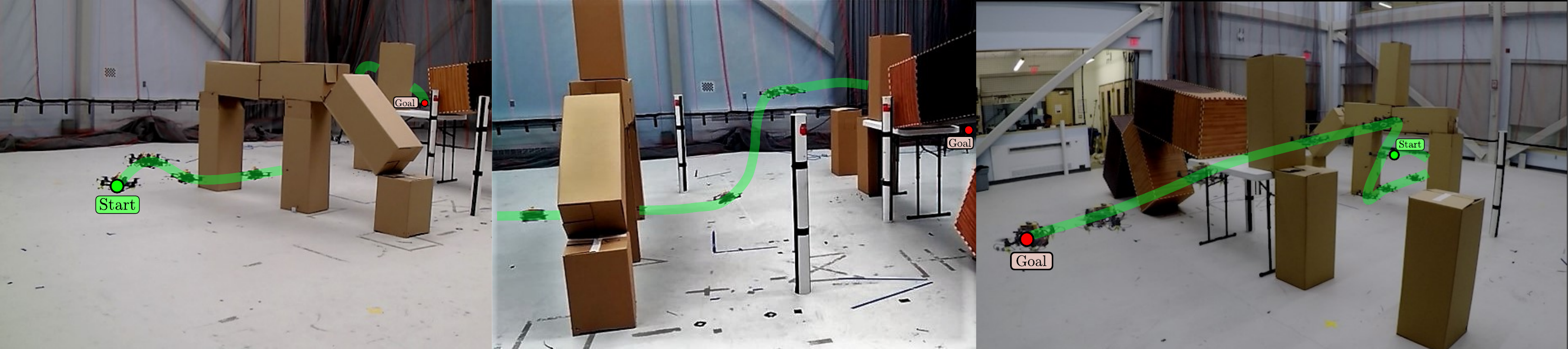}
	\caption[Composite images of Experiment~1]{Composite images of Experiment~1. The UAV must fly from start \tikzcircle[black,fill=green]{2pt} to goal \tikzcircle[black,fill=red]{2pt}. Snapshots shown every 670~ms.}
	\label{fig:exp1}
	\centering
	\includegraphics[width=\textwidth]{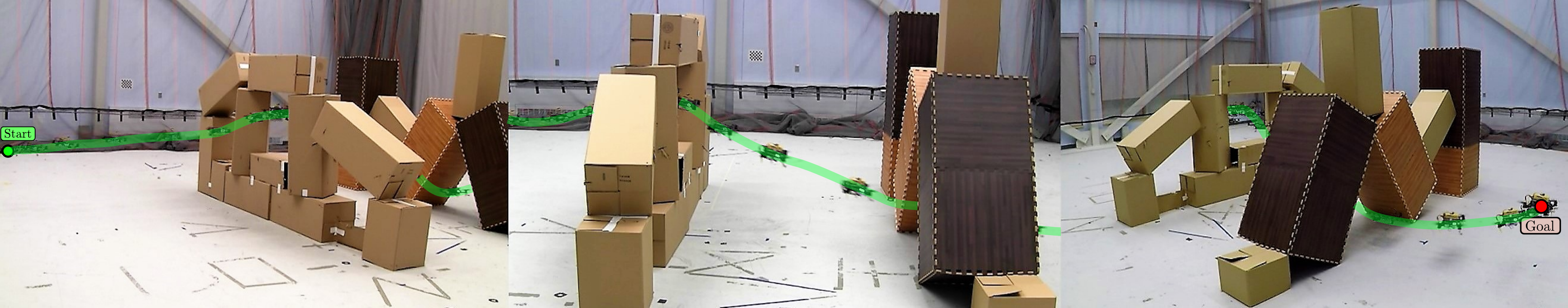}
	\caption[Composite images of Experiment~2]{ Composite image of Experiment 2. The UAV must fly from start \tikzcircle[black,fill=green]{2pt} to goal \tikzcircle[black,fill=red]{2pt}. Snapshots shown every 330~ms.}
	\label{fig:exp2}
	%
	\centering
	\includegraphics[width=\textwidth]{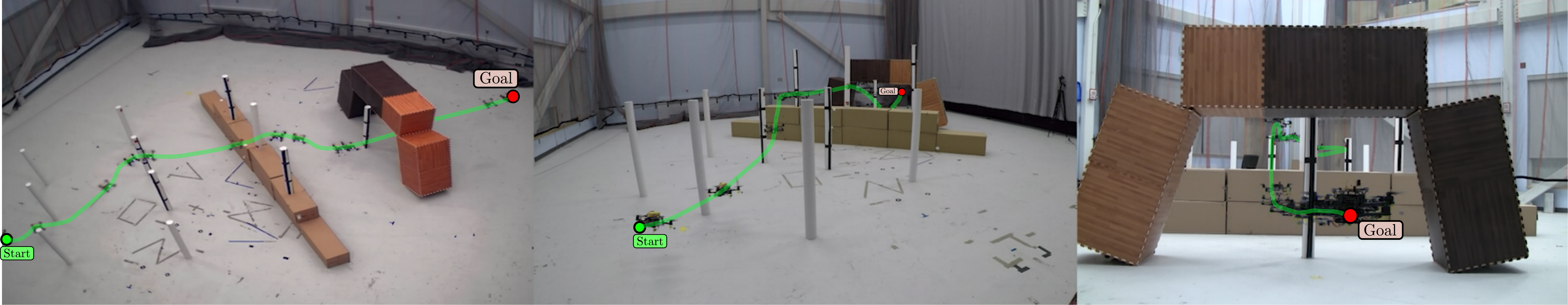}
	\caption[Composite images of Experiment~3]{ Composite image of Experiment 3. The UAV must fly from start \tikzcircle[black,fill=green]{2pt} to goal \tikzcircle[black,fill=red]{2pt}. Snapshots shown every 670~ms.}
	\label{fig:exp3}
	%
	\centering
	\includegraphics[width=\textwidth]{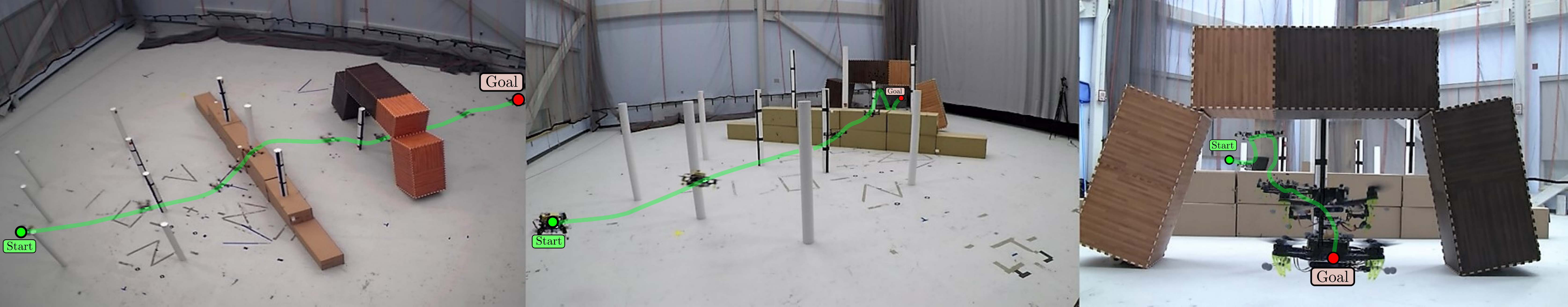}
	\caption[Composite images of Experiment~4]{ Composite image of Experiment 4. The UAV must fly from start \tikzcircle[black,fill=green]{2pt} to goal \tikzcircle[black,fill=red]{2pt}. Snapshots shown every 670~ms.}
	\label{fig:exp4}
\end{figure*} 

\subsection{Hardware}

The UAV used in the hardware experiments is shown in Fig. \ref{fig:drone}. The perception runs on the Intel\textsuperscript{\textregistered} RealSense, the mapper and planner run on the Intel\textsuperscript{\textregistered} NUC, and the control runs on the Qualcomm\textsuperscript{\textregistered} SnapDragon Flight. The attitude, IMU biases, position and velocity are estimated by fusing (via a Kalman filter) propagated IMU measurements with an external motion capture system.

The first and second experiments (Fig. \ref{fig:exp1} and \ref{fig:exp2}) were done in similar obstacle environments with the same starting point, but with different goal locations. In the first experiment (Fig. \ref{fig:exp1}), the UAV performs a 3D agile maneuver to avoid the obstacles on the table. In the second experiment (Fig. \ref{fig:exp2}) the UAV flies through the narrow gap of the cardboard boxes structure, and then flies below the triangle-shaped obstacle. In these two experiments, the maximum speed was $2.1$ m/s.

In the third and fourth experiments (Fig \ref{fig:exp3} and \ref{fig:exp4}), the UAV must fly through a space with poles of different heights, and finally below the cardboard boxes structure to reach the goal, achieving a maximum speed of $3.6$ m/s.

\begin{figure}[t]
	\centering
	\includegraphics[width=1\columnwidth]{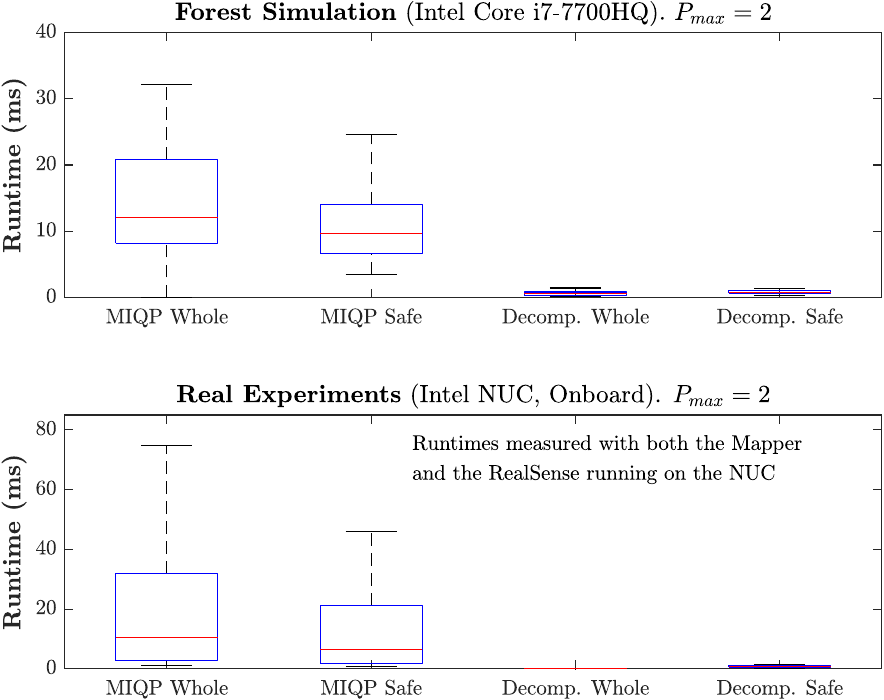}
	\caption{ Timing breakdown for the forest simulation and for the real hardware experiments. The parameters used are $P_{max}=2$, $N=10$ for the Whole Trajectory, and $N=7$ for the Safe Trajectory.}
	\label{fig:timing_real_and_sim}
\end{figure}


For $P_{max}=2$, the boxplots of the runtimes achieved on the forest simulation (measured on an Intel Core i7-7700HQ) and on the hardware experiments (measured on the onboard Intel NUC with the mapper and the RealSense also running on it) are shown in Fig. \ref{fig:timing_real_and_sim}. For the runtimes of the MIQP for the Whole and the Safe Trajectories, the 75\textsuperscript{th} percentile is always below $32$ ms.

\section{CONCLUSIONS}
\label{sec:conclusions_future_work}
This work presented FASTER, a fast and safe planner for agile flights in unknown environments. The key properties of this planner is that it leads to a higher nominal speed than other works by planning both in $\mathcal{U}$ and $\mathcal{F}$, and ensures safety by having always a Safe Trajectory planned in $\mathcal{F}$ at the beginning of every replanning step. FASTER was tested successfully both in simulated and in hardware flights, achieving velocities up to $3.6$ m/s.





\section*{ACKNOWLEDGMENT}

Thanks to Boeing Research \& Technology for support of
the hardware, to my brother Pablo Tordesillas (ETSAM-UPM) for his great help with some figures of this paper and to Parker Lusk (ACL-MIT) for his help with the hardware. The authors would also like to thank John Carter and John Ware (CSAIL-MIT) for their help with the mapper used in this paper. Supported in part by Defense Advanced Research Projects Agency (DARPA) as part of the Fast Lightweight Autonomy (FLA) program, HR0011-15-C-0110.
Views expressed here are those of the authors, and do not reflect the official views or policies of the Dept. of Defense or the U.S. Government.


\balance

\makeatletter
\def\endthebibliography{%
	\def\@noitemerr{\@latex@warning{Empty `thebibliography' environment}}%
	\endlist
}
\makeatother

\bibliographystyle{unsrt}
\bibliography{ref}

\end{document}